\documentclass[runningheads]{llncs}
\usepackage[T1]{fontenc}
\usepackage{graphicx,verbatim}
\usepackage{amsmath}
\usepackage{float}  
\usepackage{booktabs}
\usepackage{amssymb}
\usepackage{multirow}
\usepackage[table]{xcolor}

\begin{document}
\title{MedTS-TTT: Test-Time Training for\\ Medical Time Series Classification}
%

\author{
Mingzhi Chen\inst{} \and
Yiyu Gui\inst{} \and
Guibo Luo\inst{}\thanks{Corresponding author.}
}

\authorrunning{M. Chen et al.}

\institute{
Guangdong Provincial Key Laboratory of Ultra High Definition Immersive Media Technology,
Shenzhen Graduate School, Peking University, Shenzhen, China\\
\email{luogb@pku.edu.cn}
}

\maketitle              
\begin{abstract}
Medical time series (MedTS) signals such as electroencephalography (EEG) and electrocardiography (ECG) support many clinical applications. However, substantial subject-level heterogeneity often induces subject-level distribution shift, causing a fixed parameter set to generalize poorly to unseen individuals. Compared with domain adaptation methods that often depend on extra adaptation components or target-batch statistics, Test-Time Training (TTT) provides a more practical solution for sequential clinical data by enabling online adaptation from unlabeled test samples. However, many representative TTT methods require iterative inner-loop optimization, increasing test-time overhead. In this paper, we propose MedTS-TTT, a test-time training framework for medical time series modeling. MedTS-TTT is built upon Closed-Loop Self-Alignment Test-Time Training (CLSA-TTT) and a Gated Convolutional Backbone (GCB). CLSA-TTT constructs a token-level self-supervised target and performs a single-step fast-weight update for intra-layer closed-loop alignment, enabling rapid sample-wise adaptation without iterative inner-loop optimization. GCB combines CLSA-TTT-based fast adaptation and token-level fusion with a gated convolutional branch to balance local dynamic modeling and information-flow control. On 4 public datasets (2 EEG and 2 ECG) with subject-independent splits, MedTS-TTT achieves 11 top-1 rankings out of 12 evaluations across 9 baselines and 3 metrics. The code is publicly available at https://github.com/mingzhi-c/MedTS-TTT.

\keywords{Medical time series  \and Test-Time Training \and Gated Convolutional Backbone.}

\end{abstract}
\section{Introduction}

Medical time series (MedTS) data, such as electroencephalography (EEG) and electrocardiography (ECG), are widely used in clinical decision-making and continuous health monitoring. Recent deep-learning methods have shown strong performance in end-to-end classification on these signals. For example, end-to-end EEG models have been explored for neurodegenerative disorders such as Alzheimer’s disease \cite{EEGAD1,EEGAD2} and Parkinson’s disease \cite{EEGPD1,EEGPD2}. In cardiac care, ECG models have achieved strong results for arrhythmia detection \cite{ECGAD1,ECGAD2} and myocardial infarction detection \cite{ECGMI1,ECGMI2}. These advances highlight the broad potential of end-to-end MedTS modeling across clinical applications.

Recent studies have explored general-purpose backbone architectures that jointly model temporal patterns and inter-channel dependencies within a unified framework. For example, MedFormer \cite{medformer} introduces multi-granularity patching with cross-channel token interactions, MedGNN \cite{medgnn} formulates MedTS modeling as multi-resolution spatiotemporal graph learning, and MedSpaformer \cite{medspaformer} improves efficiency and transferability through dynamic token retention and redundancy compression. Despite these architectural advances, a fundamental challenge remains: MedTS exhibit substantial subject-level heterogeneity, which often leads to subject-level distribution shift, as shown in Fig. 1(a). As a result, a fixed parameter set often generalizes poorly to unseen individuals, even under the same dataset and acquisition protocol.

One line of work addresses cross-subject generalization through domain adaptation (DA), including source-free unsupervised adaptation for personalized medical time series analysis \cite{DAexample}. While effective in controlled settings, DA-based methods are often difficult to deploy in real clinical workflows because they usually introduce additional adaptation modules and frequently rely on target batches or relatively stable target statistics. In practice, clinical signals often arrive sequentially, and subject-level distribution shift may evolve over time. These constraints motivate a lightweight adaptation mechanism that can operate directly on unlabeled test samples.

Test-Time Training (TTT) provides a promising alternative by updating model parameters at inference time using self-supervised objectives on test inputs \cite{TTT1,TTT2,TTT3}. TTT is particularly suitable here because it enables online adaptation from each incoming test sample without target labels or explicit target-domain training. This makes TTT a natural fit for real-world MedTS applications, where data are often observed sequentially and adaptation must be performed under limited computational and annotation resources. However, many existing TTT methods rely on iterative inner-loop optimization to perform adaptation, which increases test-time computational overhead and complicates use in latency-sensitive scenarios. For MedTS applications, where efficient per-sample inference is often required, a more efficient TTT design is needed.

In this work, we propose MedTS-TTT, a test-time training framework for medical time series modeling. First, we introduce Closed-Loop Self-Alignment Test-Time Training (CLSA-TTT), which mitigates subject-level distribution shift during cross-subject inference by constructing a token-level self-supervised target and performing a single-step fast-weight update for intra-layer closed-loop alignment, enabling rapid sample-wise adaptation without iterative inner-loop optimization. Second, we propose a Gated Convolutional Backbone (GCB) for sequence modeling, which combines CLSA-TTT-based fast adaptation and token-level fusion with a gated convolutional branch to balance local dynamic modeling and information-flow control. We evaluate MedTS-TTT on 4 public datasets (2 EEG and 2 ECG) under subject-independent splits. MedTS-TTT achieves 11 top-1 rankings out of 12 evaluations across 9 baselines and 3 metrics, demonstrating improved robustness under subject-level distribution shift and stronger practicality in real-world settings.

\section{Method}

We propose MedTS-TTT, an end-to-end test-time training framework for medical time series classification. It is designed to address subject-level distribution shift in real-world settings without requiring extra labels or architectural changes. As shown in Fig.~1, MedTS-TTT is built upon Closed-Loop Self-Alignment Test-Time Training (CLSA-TTT) and a Gated Convolutional Backbone (GCB). Specifically, a spatiotemporal tokenizer converts multi-channel signals into a token sequence, which is then processed by the GCB, where CLSA-TTT performs a single-step self-supervised fast-weight update at test time. The final prediction is obtained by aggregating token representations.

\begin{figure}[htb]
  \centering
  \includegraphics[width=\textwidth]{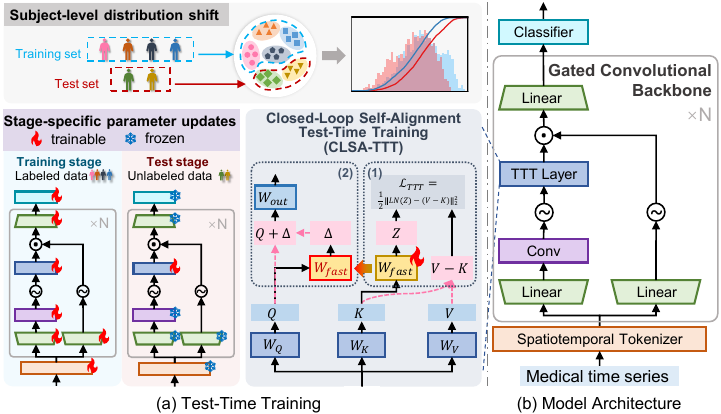}
  \caption{The comprehensive architecture of MedTS-TTT. (a) Subject-level distribution shift setting and training/test-time adaptation protocol. (b) Overall architecture of the proposed MedTS-TTT framework.}
  \label{model}
\vspace{-0.7cm}
\end{figure}

\subsection{Spatiotemporal Tokenizer for Medical Time Series}
Given a medical time series segment $X\in\mathbb{R}^{B\times C\times T}$, we design a spatiotemporal tokenizer to map it into a token sequence $H\in\mathbb{R}^{B\times L\times D}$. Specifically, we apply channel-wise temporal encoding, cross-channel spatial fusion, and patch-wise projection (with positional embeddings) to obtain $\widetilde{H}$, which serves as the input for subsequent token-level adaptation.

\textbf{Temporal encoding.} To extract shared features of short-term dynamics and spectral structure within each channel, we apply channel-wise temporal convolutions to obtain sequential representations $U=\phi(\mathcal{F}_t(X))\in\mathbb{R}^{B\times D\times C\times T}$. Here, $\phi(\cdot)$ denotes an activation function (GELU in our implementation), and $\mathcal{F}_t\in\mathbb{R}^{D\times1\times1\times k_t}$ is a channel-wise temporal convolution kernel. 

\textbf{Spatial fusion.} We fuse cross-channel information using a channel-dependent spatial convolution $\mathcal{F}_s\in\mathbb{R}^{D\times D\times C\times1}$ along the channel dimension, aggregating multi-channel signals into a compact representation $S=\phi(\mathcal{F}_s(U))\in\mathbb{R}^{B\times D\times T}$. 

\textbf{Patch embedding with positional encoding.} We partition $S$ along the temporal axis with stride $P$, and map each patch to a token via a learnable linear projection $\mathcal{P}\in\mathbb{R}^{D\times P}$, yielding a token sequence $H=\mathcal{P}(S)\in\mathbb{R}^{B\times L\times D}$, where $L=\left\lceil T\right.\left./P\right\rceil$ denotes the token sequence length. We then add a learnable positional embedding $PE\in\mathbb{R}^{L_{max}\times D}$ to preserve temporal order information:
\begin{equation}
\widetilde{H}=H+P E_{[: L]} \in \mathbb{R}^{B \times L \times D}
\end{equation}

\subsection{Closed-Loop Self-Alignment Test-Time Training}

Given a token sequence $\widetilde{H}\in\mathbb{R}^{B\times L\times D}$, we perform test-time training (TTT) inside each backbone layer by updating a small set of fast parameters with a self-supervised objective. We adopt a fast–slow parameterization: the slow parameters learn stable representation transformations, while the fast parameters $W_{fast}$ are updated in a per-sample inner loop to produce an adaptively self-aligned correction consistent with the current sequence state.

\textbf{Self-supervised objective.} We first project $\widetilde{H}$ into multi-head representations $Q,K,V\in\mathbb{R}^{B\times n\times L\times d}(d=D/n)$. TTT defines an unlabeled objective on the key–value pair by transforming $K$ with the fast parameters:
\begin{equation}
Z=KW_{fast}
\end{equation}
We then use Layer Normalization (LN) to fit, in the normalized space, the residual signal induced by $(K,V)$:
\begin{equation}
\mathcal{L}_{T T T}=\frac{1}{2}\|L N(Z)-(V-K)\|_{2}^{2}
\end{equation}
This objective defines a token-level self-supervised target, guiding the fast parameters to align with the current sample’s token structure.

\textbf{One-step fast parameter update.} To perform the inner-loop adaptation, we compute $G \triangleq \frac{\partial \mathcal{L}_{T T T}}{\partial Z} \in \mathbb{R}^{B \times n \times L \times d}$ and update the fast parameters with a single SGD step for closed-loop self-alignment:
\begin{equation}
W_{fast}\gets W_{fast}-\eta K^\top G
\end{equation}
where $\eta$ denotes a head-wise adaptive step size. After this one-step update, we apply the updated fast parameters to $Q$ to obtain an adaptive transformation:
\begin{equation}
\Delta=QW_{fast}
\end{equation}
$\Delta$ serves as the output of the adaptive representation branch at this layer and is subsequently fused with the main backbone branch. This yields one-shot, full-sequence alignment with minimal test-time overhead, making it suitable for low-latency medical time series classification.

\textbf{Training and inference usage.} During training, TTT is integrated into the forward pass as an in-layer adaptation mechanism. Given an input sample, the model first performs a one-step fast-weight update using the self-supervised objective $\mathcal{L}_{TTT}$, and then uses the updated fast weights to perform the layer forward computation and produce predictions. The outer-loop optimization is driven by the supervised cross-entropy loss $\mathcal{L}_{CE}$, which updates only the slow parameters. At test time, we retain the same in-layer TTT adaptation without label supervision. For each test sample, we apply a single fast-weight update using $\mathcal{L}_{TTT}$ and then run the forward pass for prediction, enabling adaptive inference without annotations or additional fine-tuning.

\subsection{Gated Convolutional Backbone with CLSA-TTT}

To jointly model local dynamics and enable stable adaptation at the token level, we combine a short-range convolution module $\mathcal{F}_{dw}$ with a gated feed-forward structure, and integrate a CLSA-TTT module $\mathcal{M}$ in the main branch to construct an adaptive backbone layer. Given the input token sequence of the $k\-th$ layer $\mathcal{T}^{(k)}, H^{(k)}\in\mathbb{R}^{B\times L\times D}$, the layer output is defined as
\begin{equation}
H^{(k+1)}=\mathcal{T}^{(k)}(H^{(k)})=H^{(k)}+\Delta H^{(k)}
\end{equation}
where $\mathcal{T}^{(k)}$ consists of the short-range convolution module, the CLSA-TTT module, and a gated fusion mechanism.

\textbf{Short-range convolution module.} We introduce a depthwise short-range convolution $\mathcal{F}_{dw}$ in the latent space to capture the short-term waveform variations and local rhythmic cues commonly observed in medical time series. For an input $A^{(k)}\in\mathbb{R}^{B\times L\times2D}$, the transformation is
\begin{equation}
A^{(k+1)}=\phi(\mathcal{F}_{dw}^{(k)}(A^{(k)}))\in\mathbb{R}^{B\times L\times2D}
\end{equation}

\textbf{Gate-based fusion} In the latent space, we first apply a gating projection $W_{in}\in\mathbb{R}^{D\times 4D}$ and split the result into a main branch and a gating branch:
\begin{equation}
\left[A^{(k)}, R^{(k)}\right]=H^{(k)} W_{i n}, \quad A^{(k)}, R^{(k)} \in \mathbb{R}^{B \times L \times 2 D}
\end{equation}
The main branch is processed by the short-range convolution $\mathcal{F}_{dw}^{(k)}$ to obtain $A^{(k+1)}$, which is then fed into the CLSA-TTT module $\mathcal{M}^{(k)}$ to produce a sample-dependent representation transform ${\widetilde{A}}^{(k+1)}=\mathcal{M}^{(k)}(A^{(k+1)})\in\mathbb{R}^{B\times L\times2D}$. The gating branch generates token-wise and channel-wise gating coefficients $G^{(k)}=\phi(R^{(k)})\in\mathbb{R}^{B\times L\times2D}$. We then fuse the two branches via element-wise multiplication and project back to the original dimensionality using an output projection $W_{out}\in\mathbb{R}^{2D\times D}$:
\begin{equation}
\Delta H^{(k)}=\left(G^{(k)} \odot \tilde{A}^{(k+1)}\right) W_{\text {out }} \in \mathbb{R}^{B \times L \times D}
\end{equation}
where $\bigodot$ denotes element-wise multiplication. Overall, the layer leverages short-range convolutions to model local dependencies, uses CLSA-TTT to enable one-step test-time self-alignment, and employs gating to stabilize information fusion.

\section{Experiments}

\subsection{Experimental settings}

\textbf{Datasets.} We evaluate on 4 public clinical datasets, including two EEG datasets (APAVA \cite{apava} and ADFTD \cite{adftd}) and two ECG datasets (PTB \cite{ptb} and PTB-XL \cite{ptbxl}). APAVA is used for Alzheimer’s disease detection with binary labels, while ADFTD targets differential diagnosis among healthy controls, frontotemporal dementia, and Alzheimer’s disease. For ECG, PTB provides binary labels for myocardial infarction detection, and PTB-XL supports multi-class classification of several cardiac conditions. For fair comparison and reproducibility, all datasets are processed and configured following Medformer.

\textbf{Baselines.} We compare our proposed MedTS-TTT against 7 representative state-of-the-art time-series models, including Autoformer \cite{autoformer}, Crossformer \cite{crossformer}, FEDformer \cite{fedformerfe}, Informer \cite{informer}, iTransformer \cite{itransformer}, PatchTST \cite{patchtst}, and the vanilla Transformer \cite{transformer}, as well as 2 MedTS–specific models, Medformer and MedGNN.

\textbf{Implementation details.} We employ 3 evaluation metrics: accuracy, F1 score (macro-averaged), and AUROC (macro-averaged). Model selection is based on the best validation F1 score, and the selected checkpoint is evaluated on the test set. All experiments are implemented in PyTorch and conducted on a Linux server equipped with an NVIDIA A100 GPU (80GB). For each dataset, we train the model for up to 100 epochs with an early-stopping strategy: training is terminated if the validation macro-F1 does not improve for 10 consecutive epochs, and the checkpoint achieving the best validation macro-F1 is retained for test evaluation. We optimize the model using Adam with a learning rate of $\mathrm{1\times}{\mathrm{10}}^{\mathrm{-4}}$. To ensure a consistent and fair comparison across datasets, we use an identical model configuration throughout: the patch size is set to 8, and the backbone consists of 6 stacked layers with a hidden dimension of 128.

\subsection{Results}

\subsubsection{Subject-independent Results.} We split the training, validation, and test sets at the subject level, following the same setting as Medformer \cite{medformer}. Specifically, samples from the same subject appear in only one split to prevent information leakage. This subject-independent setting reflects real-world medical time series diagnosis scenarios, where models are trained on labeled subjects and evaluated on unseen subjects to assess their disease status. In this setting, strong inter-subject variability can lead to noticeable distribution shifts, making test-time training a natural choice for improving robustness when facing new subjects.

Table 1 presents the results of the subject-independent evaluation setup. Overall, our method achieves 11 top-1 results out of 12 evaluations across 4 datasets, 9 baselines, and 3 metrics. Averaged over the four datasets, MedTS-TTT achieves the best overall performance, reaching 75.18\% in Accuracy, 70.20\% in F1 score, and 86.86\% in AUROC, which corresponds to improvements of 0.90\%, 0.65\%, and 0.83\% over the strongest baseline MedGNN, respectively. These results demonstrate that MedTS-TTT provides consistently improved robustness under subject-level distribution shift in subject-independent evaluation.

\begin{table}[ht]
\vspace{-0.2cm}
\caption{Overall Results of Subject-independent Evaluation. The best results are shown in \textbf{bold}, and the second-best results are \underline{underlined}.}
\centering
\setlength{\tabcolsep}{12pt}
\resizebox{\textwidth}{!}{%
\begin{tabular}{@{}lcccccc@{}}
\toprule
\rowcolor{teal!10}
\textbf{EEG Datasets} & \multicolumn{3}{c}{\textbf{APAVA}}            & \multicolumn{3}{c}{\textbf{ADFTD}}            \\
Models                             & Accuracy   & F1 score   & AUROC      & Accuracy   & F1 score   & AUROC      \\ \midrule
Autoformer                         & 68.64$\pm$\scriptsize{1.82} & 68.06$\pm$\scriptsize{1.94} & 75.94$\pm$\scriptsize{3.61} & 45.25$\pm$\scriptsize{1.48} & 42.59$\pm$\scriptsize{1.85} & 61.02$\pm$\scriptsize{1.82} \\
Crossformer                        & 73.77$\pm$\scriptsize{1.95} & 68.93$\pm$\scriptsize{1.85} & 72.39$\pm$\scriptsize{3.33} & 50.45$\pm$\scriptsize{2.31} & 45.50$\pm$\scriptsize{1.70} & 66.45$\pm$\scriptsize{2.03} \\
FEDformer                          & 74.94$\pm$\scriptsize{2.15} & 73.51$\pm$\scriptsize{3.39} & 83.72$\pm$\scriptsize{1.97} & 46.30$\pm$\scriptsize{0.59} & 43.91$\pm$\scriptsize{1.37} & 62.62$\pm$\scriptsize{1.75} \\
Informer                           & 73.11$\pm$\scriptsize{4.40} & 69.47$\pm$\scriptsize{5.06} & 70.46$\pm$\scriptsize{4.91} & 48.45$\pm$\scriptsize{1.96} & 45.74$\pm$\scriptsize{1.38} & 65.87$\pm$\scriptsize{1.27} \\
iTransformer                       & 74.55$\pm$\scriptsize{1.66} & 72.30$\pm$\scriptsize{1.79} & 85.59$\pm$\scriptsize{1.55} & 52.60$\pm$\scriptsize{1.59} & 46.79$\pm$\scriptsize{1.13} & 67.26$\pm$\scriptsize{1.16} \\
PatchTST                           & 67.03$\pm$\scriptsize{1.65} & 55.97$\pm$\scriptsize{3.10} & 65.65$\pm$\scriptsize{0.28} & 44.37$\pm$\scriptsize{0.95} & 41.97$\pm$\scriptsize{1.37} & 60.08$\pm$\scriptsize{1.50} \\
Transformer                        & 76.30$\pm$\scriptsize{4.72} & 73.75$\pm$\scriptsize{5.38} & 72.50$\pm$\scriptsize{6.60} & 50.47$\pm$\scriptsize{2.14} & 48.09$\pm$\scriptsize{1.59} & 67.93$\pm$\scriptsize{1.59} \\
Medformer                          & 78.74$\pm$\scriptsize{0.64} & 76.31$\pm$\scriptsize{0.71} & 83.20$\pm$\scriptsize{0.91} & 53.27$\pm$\scriptsize{1.54} & 50.65$\pm$\scriptsize{1.51} & 70.93$\pm$\scriptsize{1.19} \\
MedGNN                             & \underline{82.60$\pm$\scriptsize{0.35}} & \underline{80.25$\pm$\scriptsize{0.16}} & \underline{85.93$\pm$\scriptsize{0.26}} & \underline{56.12$\pm$\scriptsize{0.11}} & \underline{55.00$\pm$\scriptsize{0.24}} & \underline{74.68$\pm$\scriptsize{0.33}} \\
\rowcolor[HTML]{EFEFEF}
MedTS-TTT                             & \textbf{83.03$\pm$\scriptsize{1.13}} & \textbf{81.52$\pm$\scriptsize{1.45}} & \textbf{89.96$\pm$\scriptsize{1.43}} & \textbf{58.10$\pm$\scriptsize{0.82}} & \textbf{55.18$\pm$\scriptsize{2.27}} & \textbf{75.41$\pm$\scriptsize{1.25}} \\ \midrule
\rowcolor{cyan!10}
\textbf{ECG Datasets}                       & \multicolumn{3}{c}{\textbf{PTB}}              & \multicolumn{3}{c}{\textbf{PTB-XL}}           \\
Models                             & Accuracy   & F1 score   & AUROC      & Accuracy   & F1 score   & AUROC      \\ \midrule
Autoformer                         & 73.35$\pm$\scriptsize{2.10} & 63.69$\pm$\scriptsize{3.84} & 78.54$\pm$\scriptsize{3.48} & 61.68$\pm$\scriptsize{2.72} & 48.85$\pm$\scriptsize{2.27} & 82.04$\pm$\scriptsize{1.44} \\
Crossformer                        & 80.17$\pm$\scriptsize{3.79} & 72.75$\pm$\scriptsize{7.19} & 88.55$\pm$\scriptsize{3.45} & 73.30$\pm$\scriptsize{0.14} & 62.59$\pm$\scriptsize{0.14} & 90.02$\pm$\scriptsize{0.06} \\
FEDformer                          & 76.05$\pm$\scriptsize{2.54} & 67.14$\pm$\scriptsize{4.37} & 85.93$\pm$\scriptsize{4.31} & 57.20$\pm$\scriptsize{9.47} & 47.89$\pm$\scriptsize{8.44} & 82.13$\pm$\scriptsize{4.17} \\
Informer                           & 78.69$\pm$\scriptsize{1.68} & 70.84$\pm$\scriptsize{3.47} & 92.09$\pm$\scriptsize{0.53} & 71.43$\pm$\scriptsize{0.32} & 60.44$\pm$\scriptsize{0.43} & 89.65$\pm$\scriptsize{0.09} \\
iTransformer                       & 83.89$\pm$\scriptsize{0.71} & 79.06$\pm$\scriptsize{1.06} & 91.18$\pm$\scriptsize{1.16} & 69.28$\pm$\scriptsize{0.22} & 56.20$\pm$\scriptsize{0.19} & 86.71$\pm$\scriptsize{0.10} \\
PatchTST                           & 74.74$\pm$\scriptsize{1.62} & 64.36$\pm$\scriptsize{3.38} & 88.79$\pm$\scriptsize{0.91} & 73.23$\pm$\scriptsize{0.25} & \underline{62.61$\pm$\scriptsize{0.34}} & 89.74$\pm$\scriptsize{0.19} \\
Transformer                        & 77.37$\pm$\scriptsize{1.02} & 68.47$\pm$\scriptsize{2.19} & 90.08$\pm$\scriptsize{1.76} & 70.59$\pm$\scriptsize{0.44} & 59.05$\pm$\scriptsize{0.25} & 88.21$\pm$\scriptsize{0.16} \\
Medformer                          & 83.50$\pm$\scriptsize{2.01} & 79.18$\pm$\scriptsize{3.31} & \underline{92.81$\pm$\scriptsize{1.48}} & 72.87$\pm$\scriptsize{0.23} & 62.02$\pm$\scriptsize{0.37} & 89.66$\pm$\scriptsize{0.13} \\
MedGNN                             & \underline{84.53$\pm$\scriptsize{0.28}} & \underline{80.40$\pm$\scriptsize{0.62}} & \textbf{93.31$\pm$\scriptsize{0.46}} & \underline{73.87$\pm$\scriptsize{0.18}} & 62.54$\pm$\scriptsize{0.20} & \underline{90.21$\pm$\scriptsize{0.15}} \\
\rowcolor[HTML]{EFEFEF}
MedTS-TTT                             & \textbf{85.04$\pm$\scriptsize{3.12}} & \textbf{80.71$\pm$\scriptsize{4.81}} & 91.31$\pm$\scriptsize{2.65} & \textbf{74.56$\pm$\scriptsize{0.37}} & \textbf{63.39$\pm$\scriptsize{0.70}} & \textbf{90.77$\pm$\scriptsize{0.18}} \\ \bottomrule
\end{tabular}%
}
\end{table}

\begin{figure}[htb]
  \centering
  \includegraphics[width=\textwidth]{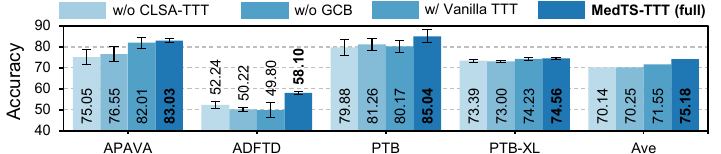}
  \caption{Performance comparison of full MedTS-TTT and ablated variants.}
  \label{ablation}
\vspace{-0.4cm}
\end{figure}

\subsubsection{Ablation Study.} To assess the roles of CLSA-TTT and GCB, we perform ablations on 4 variants: full MedTS-TTT, w/o CLSA-TTT, w/o GCB, and w/ Vanilla TTT. The w/o CLSA-TTT variant removes the CLSA-TTT module from GCB, whereas w/o GCB keeps test-time training but replaces GCB with a standard Transformer FFN block. The w/ Vanilla TTT variant replaces CLSA-TTT with a standard TTT update module \cite{TTT1}. As shown in Fig. 2, the full model performs best across all datasets. Quantitatively, replacing CLSA-TTT with Vanilla TTT reduces average accuracy by 3.63\%, supporting the effectiveness of the proposed closed-loop single-step alignment. Removing CLSA-TTT causes the largest overall degradation, with an average accuracy drop of 5.04\%, indicating that test-time training is the primary source of robustness gain under subject-level distribution shift. Removing GCB also reduces average accuracy by 4.93\%, showing that GCB provides complementary gains.

\subsubsection{Efficiency Comparison with the Vanilla TTT Mechanism.} To further evaluate CLSA-TTT against the vanilla TTT mechanism, we compare the two methods using runtime measurements. As shown in Table 2, we report mean results over 3 metrics: p50 latency, p95 latency, and throughput. Specifically, CLSA-TTT reduces p50 and p95 latency by 50.09\% and 49.76\%, respectively, and nearly doubles throughput by 99.90\%. These results show that CLSA-TTT substantially improves runtime efficiency over the vanilla TTT mechanism, making it more suitable for latency-sensitive test-time adaptation.

\begin{table}[H]
\vspace{-0.4cm}
\setlength{\tabcolsep}{6pt}
\caption{Runtime efficiency comparison between CLSA-TTT and Vanilla TTT.}
\label{table5_efficiency}
\centering
\resizebox{\columnwidth}{!}{
\begin{tabular}{cccc}
\toprule
& p50 Latency (ms) $\downarrow$ & p95 Latency (ms) $\downarrow$ & Throughput (samples per second) $\uparrow$ \\
\midrule
MedTS-TTT w/ Vanilla TTT & 29.35 & 29.70 & 4374.23 \\
\rowcolor[HTML]{EFEFEF}
MedTS-TTT w/ CLSA-TTT  & \textbf{14.65} & \textbf{14.92} & \textbf{8744.17} \\
\midrule
$\Delta$(\%)
& -50.09\%
& -49.76\%
& +99.90\% \\
\bottomrule
\end{tabular}
}
\vspace{-0.5cm}
\end{table}

\subsubsection{Effect of Test-Time Training.} To further understand why CLSA-TTT improves performance, Fig. 3 provides a qualitative analysis of subject-level distribution shift and feature alignment. Fig. 3(left) analyzes APAVA using spectral entropy, a compact measure of spectral complexity. The histogram/CDF reveals a clear subject-level distribution shift even within a single dataset (i.e., under the same acquisition center and device). Consistently, the t-SNE visualizations of features (Fig. 3(right)) show that, without CLSA-TTT, train and test features exhibit a marked distribution shift, whereas enabling CLSA-TTT substantially narrows this gap while preserving class discriminability. This qualitative observation is consistent with the quantitative gains in Fig. 2.

\begin{figure}[htb]
\vspace{-0.3cm}
  \centering
  \includegraphics[width=\textwidth]{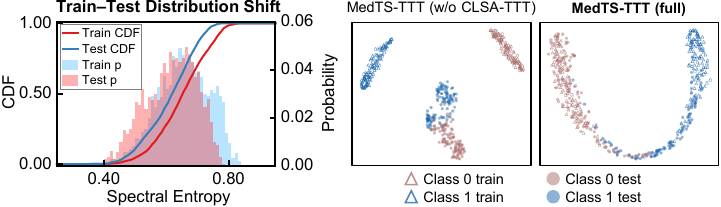}
  \caption{Distribution-Shift and Feature Alignment under Test-Time Training.}
  \label{tsne}
\vspace{-0.8cm}
\end{figure}

\section{Conclusion}

In this paper, we presented MedTS-TTT, a test-time training framework for medical time series modeling that mitigates subject-level distribution shift under cross-subject real-world settings. MedTS-TTT combines Closed-Loop Self-Alignment Test-Time Training (CLSA-TTT) for efficient single-step test-time adaptation and a Gated Convolutional Backbone (GCB) for improved token-level sequence modeling. Experiments on 4 public datasets under subject-independent splits demonstrate consistent improvements over competitive baselines. Future work will explore more efficient adaptation rules and broader real-world settings, including longer streaming sequences and larger multi-center cohorts.

%
%
%
\bibliographystyle{splncs04}
\bibliography{mybibliography}

\end{document}